\documentclass[10pt,twocolumn,letterpaper]{article}

\usepackage{cvpr}              % To produce the CAMERA-READY version
\usepackage[dvipsnames]{xcolor}

\definecolor{cvprblue}{rgb}{0.21,0.49,0.74}
\usepackage[pagebackref,breaklinks,colorlinks,citecolor=cvprblue]{hyperref}
\usepackage{booktabs}
\usepackage{multirow}
\usepackage{graphicx}
\usepackage{algorithm}
\usepackage{algpseudocode}
\usepackage{amsmath,amssymb}

\usepackage{colortbl}
\usepackage[table]{xcolor}

\definecolor{bestcolor}{RGB}{198,239,206}   % green
\definecolor{secondcolor}{RGB}{221,235,247} % blue
\definecolor{thirdcolor}{RGB}{255,235,156}  % yellow

\newcommand{\best}[1]{\cellcolor{bestcolor}\textbf{#1}}
\newcommand{\second}[1]{\cellcolor{secondcolor}#1}
\newcommand{\third}[1]{\cellcolor{thirdcolor}#1}

\usepackage{tabularx}
\title{GeoART: Geometry-Aware Refinement via Test-Time Scaling for Camera-Conditioned Video Diffusion}

\author{Liudi Yang\\
University of Freiburg\\
\and
Ziyuan Liu\\
Huawei Heisenberg Research Center (Munich)\\
}

\begin{document}

\twocolumn[{
\renewcommand\twocolumn[1][]{#1}
\maketitle
\begin{center}
    \centering
    \vspace{-15pt}
    \includegraphics[width=0.9\linewidth]{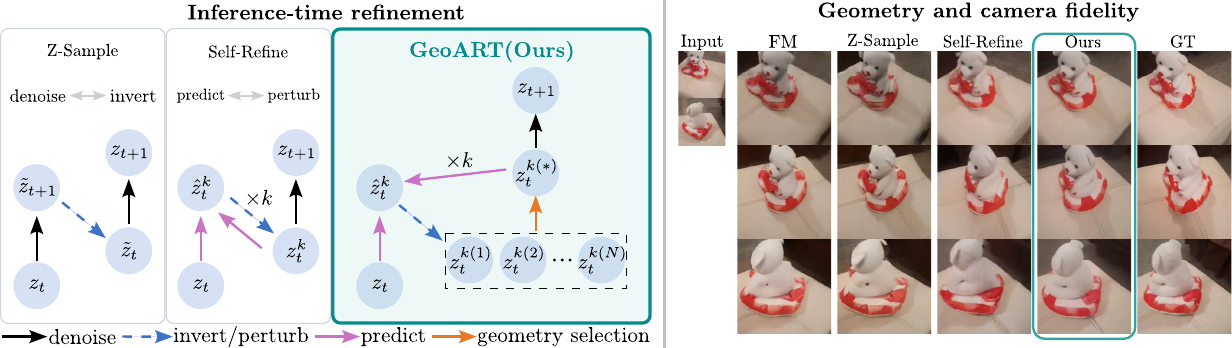}
\captionof{figure}{\textbf{Overview of GeoART.} Through inference-time self-refinement, GeoART achieves better camera fidelity and geometric consistency than standard flow-matching (FM) sampling and existing baselines~\cite{selfrefine,zsample}.}
\vspace{-5pt}
\label{fig:teaser}
\end{center}
}]
\begin{abstract}
Camera-conditioned video diffusion models often suffer from inaccurate camera control and geometric inconsistencies because camera constraints and estimated geometry are only weakly enforced during denoising. Existing remedies often require costly geometry-aware fine-tuning or additional 3D supervision. We introduce \textbf{GeoART}, a training-free self-refinement framework that improves camera controllability through geometry-aware test-time scaling. At selected denoising steps, GeoART generates multiple candidates within the Predict-and-Perturb~\cite{selfrefine} cycle and evaluates them using a geometry prior, without directly modifying the denoising trajectory. A confidence gate suppresses unreliable geometric evidence, while rank-calibrated probabilistic selection combines score magnitude and relative rank to choose the next refinement state. Experiments demonstrate consistent improvements in camera controllability, multi-view consistency, and visual quality over existing sampling and self-refinement methods, without training or modifying the pretrained video diffusion model.
\end{abstract}   
\section{Introduction}
\label{sec:intro}
Controllable video generation models have enabled camera-guided novel-view synthesis~\cite{cameractrl,motionctrl,3dgs-enhancer,ac3d,cami2v,viewcrafter}, interactive world simulation~\cite{vid2world,matrixgame,worldplay,matrix2,veo}, and embodied navigation~\cite{bar2025navigation,qin2025navigatediff,zhao2025imaginenav,zhang2024uni}. In these applications, camera trajectories or agent actions condition the generation of future observations. Despite substantial progress, existing methods still suffer from camera misalignment, geometric inconsistencies, and rendering artifacts. Even with 3D priors, denoising operates in a spatiotemporal latent space without a persistent 3D scene representation, limiting pose fidelity and cross-view consistency under large viewpoint changes and long-horizon generation.

To address these limitations, recent work has pursued two complementary directions: (1) more effectively integrating 3D priors through geometry-aware conditioning~\cite{uni3c,gen3c,confctrl,gen3r,trajectorycrafter}, and (2) optimizing pretrained models with geometry-aware rewards to improve camera alignment and geometric consistency~\cite{vggrpo,vigor,epipolardpo,li2026geo,geoflow}. Despite these advances, geometry-conditioned approaches often require large-scale training datasets with 3D supervision and remain sensitive to errors in estimated geometry, whereas reward-based approaches introduce a costly post-training stage. Inference-time self-refinement offers a training-free alternative but remains largely unexplored for improving geometric consistency~\cite{selfrefine,zsample,cfgctrl}. Existing methods refine intermediate samples by steering them toward the data manifold, improving semantic fidelity and physical plausibility. However, their lack of geometry-aware feedback limits improvements in camera controllability and geometry consistency. A natural extension is to incorporate geometric objectives through gradient-based diffusion guidance~\cite{bansal2024universal,yu2023freedom,luo2024readout,bengtson2025geometric,dong2025depthsync}. However, balancing geometric constraints with the generative prior requires carefully tuned guidance weights, while noisy or incomplete geometry can misdirect the update and degrade generation quality.

These observations motivate a different use of geometry: as an evaluation signal rather than a direct intervention in denoising. Inspired by verifier-based test-time scaling in LLMs and diffusion models~\cite{snell2024scaling,brown2024large,diffusiontts,videotts,ramesh2026test}, we extend Predict-and-Perturb self-refinement~\cite{selfrefine} by sampling multiple candidates at selected perturbation stages and evaluating their geometric consistency (Fig.~\ref{fig:teaser}). In this way, geometry guides candidate selection without directly altering the latent updates.

Based on this principle, we propose GeoART, a geometry-aware self-refinement framework for test-time scaling. GeoART scores candidate refinements using confidence-aware geometric criteria and converts these scores into a rank-calibrated sampling distribution. Rather than greedily selecting the candidate with the highest raw score, GeoART probabilistically chooses the next refinement state based on both score magnitude and relative rank, preserving exploration and reducing sensitivity to noisy evaluations. By separating geometric evaluation from latent optimization, GeoART improves camera adherence and cross-view consistency without modifying or retraining the underlying video diffusion model.

Our contributions are summarized as follows:
\begin{itemize}
\item We introduce GeoART, a training-free framework that formulates geometry-aware self-refinement as test-time scaling, using geometry priors to evaluate local refinement candidates rather than directly guide denoising.

\item We develop a confidence-gated geometry evaluator that suppresses unreliable geometric evidence when scoring candidate refinements.

\item We propose rank-calibrated probabilistic selection that balances score magnitude and relative rank, improving robustness to noisy geometry scores.

\end{itemize}

\section{Related Work}
\label{sec:related}

\subsection{Camera-Conditioned Video Diffusion}
Recent video diffusion models enable controllable viewpoint generation by conditioning on prescribed camera trajectories~\cite{motionctrl,cameractrl,cameractrl2,ac3d,recammaster,causnvs,seva,cami2v,recami2v,cat3d}. To improve cross-view consistency, subsequent methods incorporate explicit geometric cues through depth or point-cloud priors~\cite{uni3c,gen3c,confctrl,viewcrafter,trajectorycrafter,spatialcrafter,voyager}, 3D Gaussian Splatting representations~\cite{wonderland,dimensionx,3dgs-enhancer,reconx,splatflow,go2025videorfsplat}, or geometry-aware supervision~\cite{gen3r,idcnet,geometryforcing,zhang2024world}. Another line of research is geometry-aware alignment methods that improve consistency using epipolar, reconstruction, or latent 3D rewards for preference learning and reinforcement learning~\cite{epipolardpo,videogpa,camverse,worldr1,geoflow,vggrpo}. VIGOR~\cite{vigor} further employs geometric rewards for verifier-based test-time scaling. In contrast, GeoART performs geometry-aware test-time search locally within self-refinement, without post-training or modifying the underlying video diffusion model.

\subsection{Self-Refined Sampling for Diffusion Models}
Self-refinement allows a pretrained diffusion model to revisit intermediate states at inference time, typically by perturbing, inverting, or restarting denoising and regenerating from the revised state. Existing approaches use inversion--denoising cycles, restart sampling, reward-guided exploration, or uncertainty-aware perturbations to escape suboptimal trajectories and improve semantic fidelity and physical plausibility~\cite{restart,zsample,ctrlz,cfgctrl,cfgzero,selfrefine,karageorgiou2026test}. Complementary methods refine the initial noise or incorporate auxiliary-model feedback to improve temporal coherence~\cite{freeinit,videoguide}. Other approaches directly update intermediate samples using gradients from external objectives~\cite{bansal2024universal,yu2023freedom,luo2024readout}, including epipolar, depth, and multi-view constraints~\cite{bengtson2025geometric,dong2025depthsync}. However, such guidance directly alters the denoising trajectory and requires careful balancing of geometric constraints against the generative prior. GeoART instead treats geometry as an evaluator: it scores multiple self-refinement candidates and probabilistically selects the next state without gradient-based geometric intervention.

\subsection{Test-Time Scaling for Diffusion Models}
Test-time scaling improves diffusion generation by allocating additional inference computation to explore multiple samples or denoising trajectories without further training. Existing approaches use random search, particle-based resampling, or tree search to identify trajectories with higher verifier or reward scores~\cite{diffusiontts,singhal2025general,kim2025test,ramesh2026test,zhang2025classical}. EvoSearch~\cite{he2025evosearch} evolves populations of image and video trajectories through iterative selection and mutation, while Video-T1~\cite{videotts} applies verifier-guided trajectory search to video generation. However, these methods can incur substantial inference overhead and remain sensitive to verifier reliability. Hard selection and pruning are brittle under noisy scores, while commonly used perceptual and semantic rewards provide limited geometric feedback. GeoART instead performs periodic local branching within self-refinement and combines confidence-aware geometric evaluation with rank-calibrated probabilistic selection.

\begin{figure*}[t]
    \centering
    \includegraphics[width=0.9\textwidth]{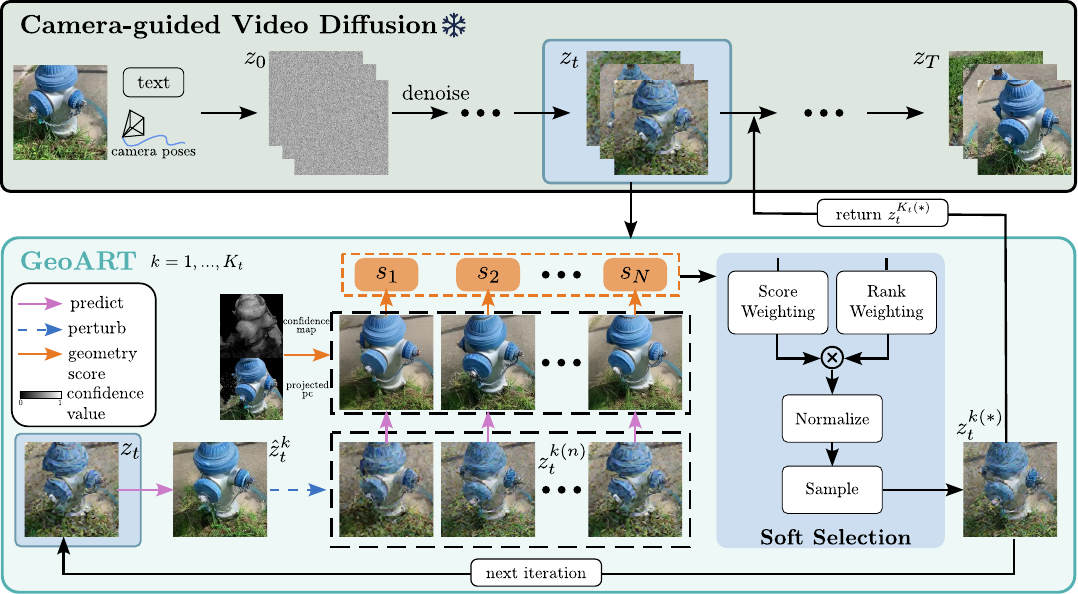}
    \caption{\textbf{GeoART framework.} At selected denoising intervals, GeoART generates multiple self-refinement candidates, evaluates them using confidence-gated geometric scores, and probabilistically selects promising candidates. After ($K_t$) refinement iterations, the refined state is returned to the original denoising trajectory without model updates or geometry-based gradient guidance.
}
    \label{fig:method}
    \vspace{-5mm}
\end{figure*}

\section{Preliminaries: Predict-and-Perturb Refinement}
\label{sec:preliminaries}

Predict-and-Perturb (P\&P)~\cite{selfrefine} is a training-free
self-refinement procedure for flow-matching models~\cite{rf,fm},
motivated by the denoising autoencoder perspective~\cite{dae}. Given a
noisy latent $z_t$, conditioning input $y$, noise level $\sigma_t$, and
model prediction $\epsilon_\theta(z_t,t,y)$, P\&P first predicts a clean
latent and then re-noises it to the current noise level. We define the
\textit{\textbf{predict}} and \textit{\textbf{perturb}} operators as
\begin{gather}
D_\theta(z_t,t,y)
=
z_t-\sigma_t\epsilon_\theta(z_t,t,y)
\label{eq:predict-operator}
\\
P_\eta(z,t)
=
(1-\sigma_t)z+\sigma_t\eta,
\qquad
\eta\sim\mathcal{N}(0,I)
\label{eq:perturb-operator}
\end{gather}
The predict operator moves the current latent toward the clean-data
manifold, while the perturb operator performs local resampling around
the prediction. Their alternation allows the model to revisit and refine
its denoising trajectory without modifying its parameters.

At solver step $t$, let $K_t$ denote the number of refinement iterations
and initialize $z_t^0=z_t$. Standard P\&P repeatedly applies the two
operators and then advances the refined latent through the original ODE
solver:
\begin{gather}
z_t^k
=
P_{\eta^k}\!\left(
D_\theta(z_t^{k-1},t,y),t
\right),
\quad
k=1,\ldots,K_t
\label{eq:pp-refinement}
\\
z_{t+1}
=
z_t^{K_t}
+
(\sigma_{t+1}-\sigma_t)
\epsilon_\theta(z_t^{K_t},t,y)
\label{eq:refined-ode-step}
\end{gather}
When $K_t=0$, we define $z_t^{K_t}=z_t$, recovering standard
flow-matching sampling. Although P\&P provides additional opportunities
for refinement, each iteration follows only one randomly perturbed
direction and does not explicitly evaluate geometric quality.

\section{Method}
\label{sec:method}

Fig.~\ref{fig:method} provides an overview of GeoART. We keep the
camera-conditioned video diffusion model frozen and refine its latent
trajectory only at inference time. The conditioning input $y$ comprises
the input views $\{I_{input}\}$, prescribed target-camera poses
$\{P_\ell^{\mathrm{cam}}\}$, and a text prompt. GeoART extends P\&P with geometry-aware local test-time scaling. At
selected solver steps, it: (1) generates multiple candidates by
independently perturbing a shared clean prediction, (2) evaluates each
candidate using a confidence-gated geometry prior, and (3)
probabilistically selects the next refinement state based on score
magnitude and relative rank. These components are detailed in
Secs.~\ref{sec:test-time-selection}, \ref{sec:geometry-scoring}, and
\ref{sec:soft-selection}.

\subsection{Geometry-Guided Test-Time Scaling}
\label{sec:test-time-selection}

A standard P\&P iteration generates only one random perturbation from
the predicted clean latent. If this perturbation leads toward an
unfavorable region, subsequent iterations may reinforce the resulting
trajectory. This limitation is particularly pronounced in camera-controlled video generation, where the denoiser primarily promotes visual quality and temporal coherence without explicitly enforcing camera-motion or cross-frame geometry constraints.

We address this limitation through local test-time search. Unlike
conventional Best-of-$N$ sampling, which generates multiple complete
videos~\cite{videotts,diffusiontts}, GeoART branches the trajectory only
at selected solver steps. All branches originate from the same clean
prediction and therefore preserve similar semantic and appearance content while exploring different local refinement directions.

Let $K_t=\mathcal{R}(t)$ denote the number of refinement iterations at
step $t$, where $K_t=0$ skips refinement, and initialize $z_t^0=z_t$.
At refinement iteration $k$, GeoART evaluates the frozen denoiser at the
current state, obtaining
$\epsilon^k=\epsilon_\theta(z_t^{k-1},t,y)$.
Using this prediction, it estimates the corresponding clean latent:
\begin{equation}
\hat z_t^k
=
z_t^{k-1}-\sigma_t\epsilon^k
\label{eq:clean-prediction}
\end{equation}

At selection steps satisfying
$t\equiv 0\pmod{\Delta}$, where $\Delta$ denotes the
selection interval in solver steps, GeoART branches the refinement
trajectory by independently re-noising the shared clean prediction
$\hat z_t^k$. Specifically, it generates $N$ candidates:
\begin{equation}
z_t^{k(i)}
=
(1-\sigma_t)\hat z_t^k+\sigma_t\eta_i,
\quad
\eta_i\sim\mathcal{N}(0,I)
\label{eq:candidate-generation}
\end{equation}
Because the candidates differ only in their sampled perturbations, they
provide alternative local directions around the same clean prediction.
Each candidate is denoised once more to obtain a look-ahead clean
prediction for geometry-aware evaluation. The resulting scores define a
soft selection distribution from which we sample $i^\star$ and update
the refinement state as $z_t^k=z_t^{k(i^\star)}$. At non-selection steps, GeoART retains the standard single-perturbation P\&P update. 
% A non-selection refinement iteration requires one
% additional denoiser evaluation, whereas a selection iteration requires
% one shared prediction and $N$ candidate look-ahead evaluations. Thus,
% GeoART allocates additional computation only at selected parts of the
% trajectory instead of generating $N$ complete videos. The refinement
% schedule $\mathcal{R}$, selection interval $\Delta$, and candidate count
% $N$ provide direct controls over the inference-quality trade-off.

\subsection{Confidence-Gated Geometry Scoring}
\label{sec:geometry-scoring}

Reliable candidate evaluation requires a geometry prior aligned with
the target video trajectory. However, off-the-shelf geometry estimators
can be inaccurate in occluded or textureless regions.
Treating all geometric estimates equally may therefore corrupt the score
and lead to unreliable candidate selection.

\noindent\textbf{Geometry-prior construction.}
Before diffusion sampling, VGGT~\cite{vggt} processes the conditioning
views $\{I_{input}\}$ to reconstruct a scene point cloud and estimate
corresponding point-wise confidence. We project the reconstructed point
cloud and its confidence to the prescribed target-camera poses
$\{P_\ell^{\mathrm{cam}}\}$, producing a geometry-projected video and a
spatially aligned confidence video. Pixels without a valid point
projection, including missing and disoccluded regions, are assigned zero
confidence.

The projected point-cloud video is encoded once using the frozen video
VAE~\cite{vae}, producing the geometry latent $g$. The confidence video
is linearly interpolated to the spatial and temporal resolution of $g$
and broadcast across the latent channels to obtain $c$. Pixels without valid projections are set to zero. Consequently,
$g$, $c$, and the diffusion latent share the same dimensions. Both $g$ and $c$ are computed once and reused throughout sampling.

\noindent\textbf{Candidate evaluation.}
We construct a confidence gate that removes unreliable regions while
weighting retained elements according to their confidence:
\begin{equation}
w_j
=
\begin{cases}
c_j, & c_j\geq\tau_c,\\
0,   & c_j<\tau_c,
\end{cases}
\label{eq:confidence-gate}
\end{equation}
where $j$ indexes the spatiotemporal latent elements and $\tau_c$ is the
confidence threshold. Thus, low-confidence elements are discarded,
while reliable elements contribute proportionally to their confidence. For each re-noised candidate $z_t^{k(i)}$, we perform one look-ahead
clean-latent prediction:
\begin{equation*}
\hat z_t^{k(i)}
=
z_t^{k(i)}
-
\sigma_t\epsilon_\theta(z_t^{k(i)},t,y)
\end{equation*}
We then compare this prediction with the geometry latent $g$ using
\begin{equation}
s^{k(i)}
=
-
\operatorname{mean}
\left[
w\odot
\left(
\hat z_t^{k(i)}-g
\right)^2
\right]
\label{eq:geometry-score}
\end{equation}
A higher score indicates stronger agreement with reliable regions of the
geometry prior. 
% The same gate $w$ is used for every candidate associated with a given
% input, ensuring that candidates are compared over identical geometric regions. If no element passes the confidence threshold, all candidates
% receive the same score and selection reduces to uniform sampling.
% Because all quantities are cached and aligned in the VAE latent space,
% candidate evaluation requires one additional denoiser call but neither
% VAE decoding nor additional VGGT inference.

\subsection{Rank-Calibrated Soft Selection}
\label{sec:soft-selection}

Given $N$ candidate scores, a direct strategy is to retain the candidate
with the highest score. However, this hard Best-of-$N$ decision can
follow an incorrect refinement direction when the geometry estimates
are noisy or candidate scores are similar. Applying a softmax directly
to the raw scores preserves exploration but remains sensitive to
changes in score scale across solver steps and scenes.

We therefore combine absolute score magnitude with relative rank. Let
$S$ denote a random candidate score with cumulative distribution
function $F(s)=\Pr(S\leq s)$. For a candidate with score $s$, the
probability of outperforming $N-1$ independent competitors is
$F(s)^{N-1}$. This order-statistic interpretation motivates weighting
each candidate according to its empirical percentile within the current
candidate set. Because $F$ is unknown, we estimate it using the self-inclusive
empirical CDF and combine the \textbf{rank weighting} with exponential \textbf{score weighting}:
\begin{equation}
\begin{aligned}
\widehat F^{k(i)}
&=
\frac{1}{N}
\sum_{j=1}^{N}
\mathbf{1}
\left\{
s^{k(j)}\leq s^{k(i)}
\right\},\\
p^{k(i)}
&=
\frac{
\exp\left(\lambda s^{k(i)}\right)
\left(\widehat F^{k(i)}\right)^{N-1}
}{
\sum_{j=1}^{N}
\exp\left(\lambda s^{k(j)}\right)
\left(\widehat F^{k(j)}\right)^{N-1}
}.
\end{aligned}
\label{eq:soft-selection}
\end{equation}
The empirical CDF measures the fraction of candidates whose scores do
not exceed that of candidate $i$, providing scale-invariant relative
calibration. The exponential term retains sensitivity to score
magnitude, while the rank term favors candidates that consistently
outperform their competitors. We set $\lambda=3$ in all experiments.

Finally, we sample
$i^\star\sim\operatorname{Categorical}(p^k)$ and update
$z_t^k=z_t^{k(i^\star)}$. This preserves local exploration and avoids
deterministically following a potentially noisy maximum score. The complete procedure is summarized in Algorithm~\ref{alg:geoart}.
\begin{algorithm}[hbpt]
\caption{GeoART Sampling}
\label{alg:geoart}
\footnotesize
\begin{algorithmic}[1]
\Require Condition $y$, geometry latent $g$, confidence $c$,
noise schedule $\{\sigma_t\}$, refinement schedule $\mathcal R$,
interval $\Delta$, candidates $N$
\Ensure Generated video $V$
\State $z_0\sim\mathcal N(0,I)$;\quad
$w\gets\operatorname{ConfidenceGate}(c)$
\For{$t=0,\ldots,T-1$}
    \State $K_t\gets\mathcal R(t)$;\quad $z_t^{0}\gets z_t$
    \For{$k=1,\ldots,K_t$}
        \State $\epsilon^{k}
        \gets\epsilon_\theta(z_t^{k-1},t,y)$;\quad
        $\hat z_t^{k}
        \gets z_t^{k-1}-\sigma_t\epsilon^{k}$
        \If{$t\bmod\Delta=0$}
            \For{$i=1,\ldots,N$}
                \State ${z}_t^{k(i)}
                \gets(1-\sigma_t)\hat z_t^{k}+\sigma_t\eta_i$,
                \quad $\eta_i\sim\mathcal N(0,I)$
                \State $\hat z_t^{k(i)}
                \gets{z}_t^{k(i)}
                -\sigma_t\epsilon_\theta({z}_t^{k(i)},t,y)$
                \State $s^{k(i)}
                \gets-\operatorname{mean}
                [w\odot(\hat z_t^{k(i)}-g)^2]$
            \EndFor
            \State $p^{k}
            \gets\operatorname{RankCalibratedDistribution}(\{s^{k(i)}\}_{i=1}^{N})$
            \State $i^\star\sim\operatorname{Categorical}(p^{k})$;
            \quad $z_t^{k}\gets z_t^{k(i^\star)}$
        \Else
            \State $z_t^{k}
            \gets(1-\sigma_t)\hat z_t^{k}+\sigma_t\eta$,
            \quad $\eta\sim\mathcal N(0,I)$
        \EndIf
    \EndFor
    \State $\epsilon\gets\epsilon_\theta(z_t^{K_t},t,y)$;\quad
    $z_{t+1}\gets
    z_t^{K_t}+(\sigma_{t+1}-\sigma_t)\epsilon$
\EndFor
\State \Return $V\gets\operatorname{Decode}_{\mathrm{VAE}}(z_T)$
\end{algorithmic}
\end{algorithm}
\vspace{-15pt}

\section{Experiment}
\label{sec:experiment}
\begin{figure*}[t]
    \centering
    \includegraphics[width=0.9\textwidth]{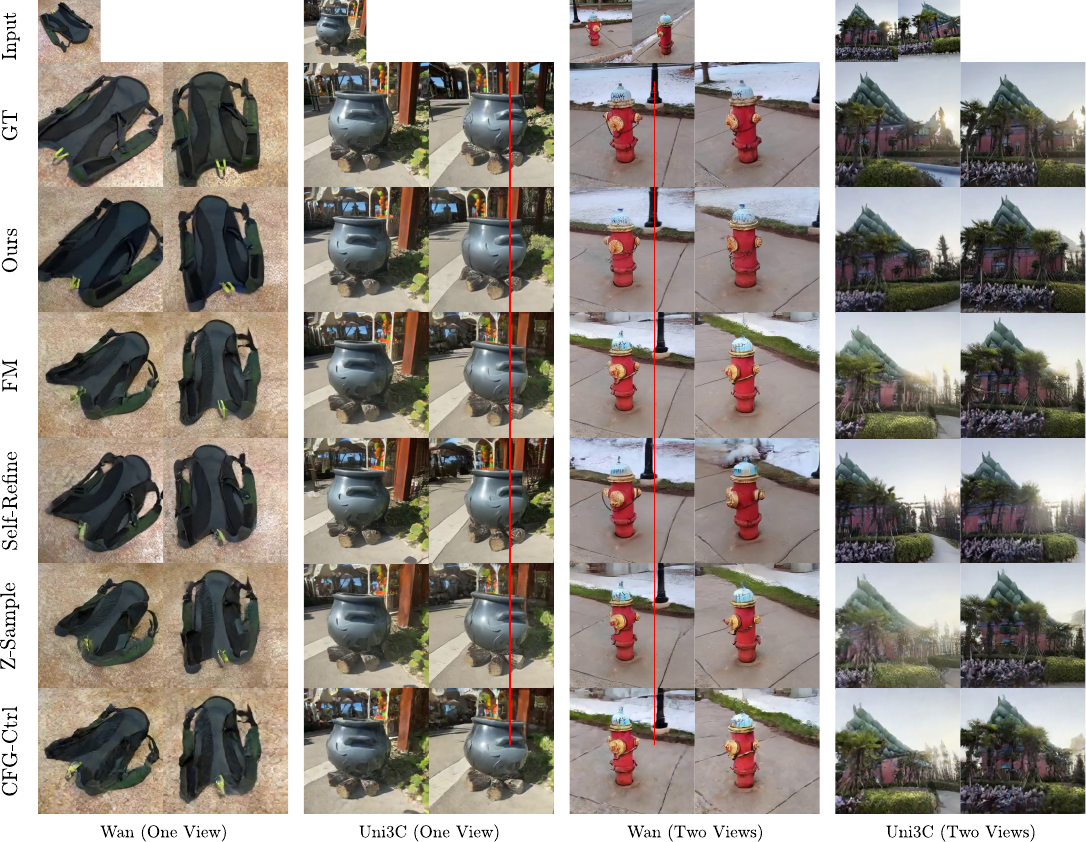}
    \caption{\textbf{Comparison with baseline methods.}
    Our method achieves higher visual quality, fewer artifacts, and more accurate camera control. \textcolor{red}{Red} lines mark the ground-truth positions of selected vertical structures (e.g., columns and poles) and remain fixed across methods to visualize positional drift.}
    \label{fig:baseline-comparison}
    \vspace{-3mm}
\end{figure*}

\noindent\textbf{Implementation details.}
We evaluate GeoART on two camera-conditioned video diffusion backbones, Wan2.1-Fun-Control-Camera~\cite{wan} and Uni3C~\cite{uni3c}, to assess its generalizability across architectures. Both backbones are fine-tuned on a combined training set of 800 sequences: 200 from DL3DV~\cite{dl3dv} and 600 from CO3D~\cite{co3d}. All frames are resized to $256\times256$, and 33 frames are uniformly sampled from each sequence to form a training clip. Each backbone is fine-tuned for 20K iterations, requiring approximately 24 hours on a single GPU. GeoART is subsequently applied without further training or model modification. 

\noindent\textbf{Evaluation protocol.}
We consider two levels of input sparsity: single-view conditioning and two-view head-and-tail conditioning, where the first and last views are provided to the video diffusion model. We evaluate all methods on 100 held-out scenes from CO3D and DL3DV using five random seeds per scene. All metrics are averaged over the $500$ generations. Each inference run requires approximately 30 GB of GPU memory. Following Self-Refine~\cite{selfrefine}, the refinement schedule is $\mathcal{R}=\{(5,15,2),(40,45,1)\}$, where each tuple specifies the start step, end step, and number of refinement iterations, respectively. Additional implementation details are provided in the supplementary material.

% Caption-safe wrapper for cell-color ranking macros.
\DeclareRobustCommand{\ranklegend}[1]{%
  \begingroup
  \setlength{\tabcolsep}{2pt}%
  \renewcommand{\arraystretch}{0.9}%
  \begin{tabular}[c]{c}#1\end{tabular}%
  \endgroup
}

\begin{table*}[t]
\centering
\setlength{\tabcolsep}{3pt}
\renewcommand{\arraystretch}{1.1}
\small
\resizebox{0.95\textwidth}{!}{%
\begin{tabular}{cllcccccccccc}
\toprule
\textbf{Model}
& \textbf{Views}
& \textbf{Method}
& \textbf{NFE} $\downarrow$
& \textbf{PSNR} $\uparrow$
& \textbf{SSIM} $\uparrow$
& \textbf{LPIPS} $\downarrow$
& \textbf{Trans.} $\downarrow$
& \textbf{Rot.} $\downarrow$
& \textbf{FID} $\downarrow$
& \textbf{FVD} $\downarrow$
& \textbf{BG Cons.} $\uparrow$
& \textbf{Subj. Cons.} $\uparrow$ \\
\midrule

% Wan: one-view
\multirow{18}{*}{Wan}
& \multirow{9}{*}{One}
& FM
& 50
& 15.23 & 0.409 & 0.415
& 0.332 & 0.247 & 109.43 & 108.47
& 0.908 & 0.805 \\

& & FM ($2\times$ NFE)
& 100
& 15.12 & 0.404 & \second{0.403}
& 0.297 & 0.225 & 105.83 & 102.16
& \third{0.917} & 0.807 \\

& & Self-Refine~\cite{selfrefine}
& 78
& 15.26 & \third{0.425} & \best{0.399}
& \second{0.285} & \third{0.210}
& 109.28 & \second{88.53}
& \best{0.919} & \third{0.822} \\

& & CFG-Ctrl~\cite{cfgctrl}
& 50
& 15.19 & 0.406 & 0.408
& 0.335 & 0.266 & 127.89 & 140.85
& 0.911 & 0.803 \\

& & Z-Sample~\cite{zsample}
& 130
& 14.96 & 0.392 & 0.425
& 0.351 & 0.263 & 110.86 & 98.87
& 0.914 & 0.821 \\

& & Video-T1~\cite{videotts}
& 500
& \second{15.40} & 0.411 & 0.405
& 0.310 & 0.225 & \third{105.65} & 104.33
& \second{0.918} & \best{0.826} \\

& & VIGOR~\cite{vigor}
& 500
& \third{15.38} & \second{0.428} & \third{0.404}
& \third{0.289} & \second{0.208}
& 107.22 & \third{92.76}
& \second{0.918} & \second{0.825} \\

& & GC-Ref~\cite{bengtson2025geometric}
& 350
& 15.26 & 0.395 & 0.409
& 0.298 & 0.254 & \second{105.26} & 94.23
& 0.905 & 0.812 \\

& & \textbf{Ours}
& 118
& \best{16.61} & \best{0.456} & \best{0.399}
& \best{0.280} & \best{0.203}
& \best{104.77} & \best{84.68}
& 0.913 & \second{0.825} \\

\cmidrule(lr){2-13}

% Wan: two-view
& \multirow{9}{*}{Two}
& FM
& 50
& 16.53 & 0.429 & 0.334
& 0.251 & 0.189
& 101.34 & 106.90
& 0.920 & \third{0.834} \\

& & FM ($2\times$ NFE)
& 100
& 16.61 & 0.444 & 0.333
& 0.248 & 0.177
& 97.10 & 92.17
& 0.919 & \second{0.836} \\

& & Self-Refine~\cite{selfrefine}
& 78
& \third{17.01} & \third{0.446} & \second{0.320}
& \third{0.187} & \third{0.137}
& 91.83 & 87.29
& \third{0.923} & 0.832 \\

& & CFG-Ctrl~\cite{cfgctrl}
& 50
& 16.94 & \second{0.453} & 0.336
& 0.287 & 0.219
& 116.53 & 145.68
& 0.915 & 0.808 \\

& & Z-Sample~\cite{zsample}
& 130
& 16.88 & 0.432 & 0.336
& 0.211 & 0.161
& 89.47 & 88.78
& 0.914 & 0.831 \\

& & Video-T1~\cite{videotts}
& 500
& \second{17.08} & 0.442 & 0.325
& 0.196 & 0.141
& \second{87.57} & \third{79.80}
& \third{0.923} & \third{0.834} \\

& & VIGOR~\cite{vigor}
& 500
& 16.97 & 0.444 & \third{0.322}
& \second{0.180} & \second{0.128}
& \third{87.68} & \second{77.97}
& \best{0.926} & \second{0.836} \\

& & GC-Ref~\cite{bengtson2025geometric}
& 350
& 16.85 & 0.435 & 0.343
& 0.208 & 0.139
& 88.39 & 92.43
& 0.922 & 0.827 \\

& & \textbf{Ours}
& 118
& \best{17.60} & \best{0.475} & \best{0.318}
& \best{0.151} & \best{0.109}
& \best{82.22} & \best{76.05}
& \second{0.924} & \best{0.837} \\

\midrule

% Uni3C: one-view
\multirow{18}{*}{Uni3C}
& \multirow{9}{*}{One}
& FM
& 50
& 15.35 & 0.402 & 0.406
& 0.318 & 0.232
& 106.09 & 100.14
& 0.914 & 0.803 \\

& & FM ($2\times$ NFE)
& 100
& 15.24 & 0.407 & 0.405
& 0.312 & 0.236
& 100.93 & 95.21
& 0.918 & 0.812 \\

& & Self-Refine~\cite{selfrefine}
& 78
& \third{16.02} & \third{0.436} & \second{0.346}
& \best{0.272} & \best{0.195}
& 96.25 & \second{89.64}
& \third{0.921} & \third{0.829} \\

& & CFG-Ctrl~\cite{cfgctrl}
& 50
& 15.97 & 0.411 & \third{0.360}
& 0.290 & 0.219
& 116.89 & 140.21
& 0.913 & 0.816 \\

& & Z-Sample~\cite{zsample}
& 130
& 15.88 & 0.412 & 0.370
& \third{0.279} & 0.217
& 94.69 & 106.23
& 0.919 & 0.826 \\

& & Video-T1~\cite{videotts}
& 500
& \second{16.05} & 0.423 & \third{0.360}
& 0.280 & 0.203
& \second{91.95} & \third{90.66}
& \second{0.922} & \best{0.832} \\

& & VIGOR~\cite{vigor}
& 500
& 16.00 & 0.422 & 0.361
& 0.286 & 0.206
& \third{92.61} & 98.24
& \second{0.922} & \third{0.829} \\

& & GC-Ref~\cite{bengtson2025geometric}
& 350
& 15.93 & \second{0.442} & 0.372
& 0.309 & \third{0.202}
& 99.56 & 102.44
& 0.919 & \second{0.830} \\

& & \textbf{Ours}
& 118
& \best{16.91} & \best{0.449} & \best{0.328}
& \second{0.273} & \second{0.200}
& \best{89.50} & \best{81.02}
& \best{0.925} & \best{0.832} \\

\cmidrule(lr){2-13}

% Uni3C: two-view
& \multirow{9}{*}{Two}
& FM
& 50
& 16.61 & \second{0.441} & 0.351
& 0.251 & 0.192
& 104.41 & 104.09
& 0.921 & 0.813 \\

& & FM ($2\times$ NFE)
& 100
& 16.50 & \third{0.436} & 0.340
& 0.249 & 0.193
& 100.20 & 98.01
& 0.922 & 0.825 \\

& & Self-Refine~\cite{selfrefine}
& 71
& \second{16.74} & \second{0.441} & \second{0.332}
& 0.254 & 0.201
& \third{91.53} & 96.81
& \best{0.927} & \third{0.834} \\

& & CFG-Ctrl~\cite{cfgctrl}
& 50
& 16.19 & 0.418 & 0.350
& \third{0.229} & \third{0.164}
& 115.81 & 132.36
& 0.917 & 0.806 \\

& & Z-Sample~\cite{zsample}
& 130
& 16.03 & 0.417 & 0.363
& 0.243 & 0.181
& 92.38 & 92.05
& 0.922 & 0.830 \\

& & Video-T1~\cite{videotts}
& 500
& 16.39 & 0.433 & 0.343
& \second{0.217} & \second{0.158}
& \second{91.42} & \second{77.73}
& \third{0.923} & \second{0.835} \\

& & VIGOR~\cite{vigor}
& 500
& 16.60 & 0.429 & 0.339
& 0.230 & 0.167
& 95.21 & \third{88.62}
& 0.919 & 0.823 \\

& & GC-Ref~\cite{bengtson2025geometric}
& 350
& \third{16.64} & 0.430 & \third{0.335}
& 0.241 & 0.175
& 95.80 & 102.36
& 0.914 & 0.822 \\

& & \textbf{Ours}
& 118
& \best{17.32} & \best{0.462} & \best{0.319}
& \best{0.184} & \best{0.137}
& \best{90.42} & \best{74.31}
& \second{0.925} & \best{0.836} \\

\bottomrule
\end{tabular}%
}
\caption{Quantitative comparison using Wan and Uni3C under one-view and
two-view settings. Best, second-best, and third-best values within each
model/view setting are highlighted in
\ranklegend{\best{\textbf{green}}},
\ranklegend{\second{\textbf{blue}}} and
\ranklegend{\third{\textbf{yellow}}}, respectively.}
\label{tab:quantitative_comparison}
\vspace{-1em}
\end{table*}

\noindent{\textbf{Evaluation Metrics.}}
We first evaluate novel-view reconstruction fidelity using PSNR, SSIM,
and LPIPS. We then measure camera controllability using translation error
(\emph{$E_T$}) and rotation error (\emph{$E_R$}) with respect to
the target trajectory. Let $(\mathbf{R}_i,\mathbf{t}_i)$ and
$(\hat{\mathbf{R}}_i,\hat{\mathbf{t}}_i)$ denote the target pose and the aligned
pose estimated by Pi3~\cite{pi3} and Umeyama algorithm~\cite{Umeyama} at frame $i$, respectively. The two
errors are computed as
\begin{alignat}{2}
&E_T
&&=
\frac{1}{M}\sum_{i=1}^{M}
\left\lVert
\hat{\mathbf{t}}_i-\mathbf{t}_i
\right\rVert_2
\\
&E_R
&&=
\frac{1}{M}\sum_{i=1}^{M}
\arccos\left(
\frac{
\operatorname{tr}
\left(\hat{\mathbf{R}}_i^{\top}\mathbf{R}_i\right)-1
}{2}
\right)
\end{alignat}
where $M$ is the number of evaluated frames. We further report FID and
FVD to assess image- and video-level generative quality, respectively. Finally, we use the
subject-consistency and background-consistency metrics from
VBench~\cite{vbench,vbench++} to evaluate temporal consistency across
frames.

\subsection{Comparison with baseline}
\noindent \textbf{Baselines}
We compare our method with alternative inference-time sampling and refinement strategies under the experimental setup described above. For each backbone and conditioning setting, we evaluate the standard flow-matching (FM) sampler of the corresponding base model and a higher-compute variant using twice the number of function evaluations (FM \(2\times\) NFE). We further include three training-free sampling-refinement methods: Self-Refine~\cite{selfrefine}, Z-Sample~\cite{zsample}, and CFG-Ctrl~\cite{cfgctrl}, as well as the geometry-guided test-time refinement method GC-Ref~\cite{bengtson2025geometric}. Finally, we compare against two video test-time scaling methods,
Video-T1~\cite{videotts} and VIGOR~\cite{vigor}, using a Best-of-$10$
protocol that generates ten candidates per input and retains the
highest-scoring output. All baselines use identical model weights, conditioning inputs, input views, and evaluation protocols. We report the effective NFE of each method to make differences in inference-time computation explicit.

Table~\ref{tab:quantitative_comparison} summarizes the quantitative results. GeoART achieves the strongest overall performance across both backbones and conditioning settings. It consistently obtains the best PSNR, SSIM, LPIPS, FID, and FVD scores, while also substantially reducing translation and rotation errors in three of the four settings. Under Uni3C single-view conditioning, its pose accuracy remains competitive, ranking second in both translation and rotation error. GeoART also achieves strong background and subject consistency, demonstrating that improved camera adherence does not compromise temporal coherence.

Figure~\ref{fig:baseline-comparison} presents qualitative comparisons across both backbones and conditioning settings. GeoART more faithfully preserves object geometry and scene structure while following the prescribed camera trajectory. In particular, its outputs remain closely aligned with the ground truth and the red reference lines, whereas competing methods exhibit positional drift, geometric deformation, and inconsistent object scale. GeoART also produces sharper object boundaries and fewer rendering artifacts, demonstrating that its improved camera fidelity does not compromise visual quality. Although GeoART requires additional denoiser evaluations, its consistent improvements demonstrate a favorable quality--computation trade-off.

\subsection{Hyper-parameter Analysis}
\label{sec:hyperparameter}
\begin{figure}[t]
\vspace{-2pt}
\centering
\includegraphics[width=0.92\linewidth]{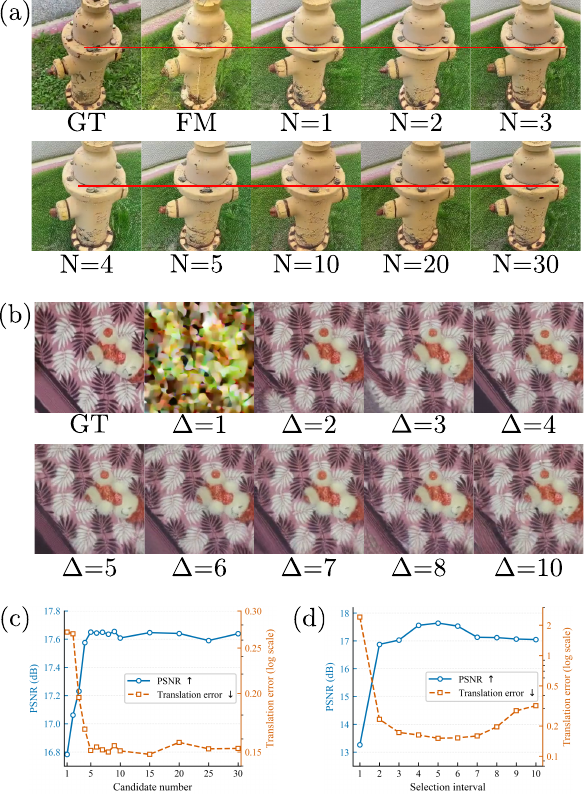}
\caption{\textbf{Hyperparameter analysis.}
Qualitative and quantitative effects of the candidate number $N$ (a,c)
and selection interval $\Delta$ (b,d). The red line in (a) marks the ground-truth object position
and remains fixed across settings.}
\label{fig:candidate-num}
\vspace{-2em}
\end{figure}

\noindent\textbf{Candidate number $N$.}
We first investigate the effect of the number of candidates sampled at each perturbation step. As shown in Fig.~\ref{fig:candidate-num}(a,c), increasing the candidate pool initially improves geometric consistency and camera alignment by providing more diverse refinement directions. However, the gains gradually saturate and exhibit minor fluctuations, while the inference cost grows with the number of candidates. We therefore use five candidates by default, achieving a favorable balance between generation quality and computational efficiency.

\noindent\textbf{Selection interval $\Delta$.}
As shown in Fig.~\ref{fig:candidate-num}(b,d), performance initially improves as the selection interval increases, but begins to degrade when the interval becomes excessively large. Short intervals invoke candidate selection too frequently, potentially causing oscillations in the sampling trajectory and introducing refinement artifacts. Conversely, long intervals provide insufficient opportunities for test-time correction, allowing errors to accumulate and making them harder to rectify. These results highlight a trade-off between correction frequency and sampling stability, with a moderate selection interval achieving the best reconstruction quality and pose accuracy. Additional hyperparameter analyses are provided in the supplementary material.
\subsection{Ablation Study}
\label{sec:ablation}
\begin{table}[t]
\centering
\setlength{\tabcolsep}{3pt}
\renewcommand{\arraystretch}{1.08}
\scriptsize
\resizebox{0.95\columnwidth}{!}{%
\begin{tabular}{@{}lccccc@{}}
\toprule
\textbf{Configuration}
& \textbf{PSNR} $\uparrow$
& \textbf{SSIM} $\uparrow$
& \textbf{LPIPS} $\downarrow$
& \textbf{Trans.} $\downarrow$
& \textbf{Rot.} $\downarrow$ \\
\midrule

\multicolumn{6}{@{}l@{}}{\textit{Component and selection ablations}} \\
Single candidate ($N=1$)
& 17.008 & 0.429 & 0.393 & 0.264 & 0.192 \\
Uniform selection
& 17.424 & 0.435 & 0.396 & 0.236 & 0.173 \\
Greedy selection
& 17.594 & 0.433 & 0.384 & 0.219 & 0.194 \\
w/o rank calibration
& 17.602 & 0.461 & 0.357 & 0.228 & 0.172 \\
w/o confidence gate
& 17.370 & 0.446 & 0.363 & 0.237 & 0.189 \\

\midrule
\multicolumn{6}{@{}l@{}}{\textit{Geometry evaluators}} \\
Ours (Pi3)~\cite{pi3}
& 17.762 & \textbf{0.476} & 0.335
& \textbf{0.202} & 0.151 \\
Ours (VGGT-$\Omega$)~\cite{vggtomega}
& 17.754 & 0.475 & 0.341
& 0.206 & 0.158 \\
Ours (MASt3R)~\cite{mast3r}
& 17.763 & 0.472 & \textbf{0.333}
& 0.208 & 0.150 \\
Ours (DUSt3R)~\cite{dust3r}
& 17.736 & 0.472 & 0.336
& 0.215 & 0.156 \\
\textbf{Ours (VGGT)}~\cite{vggt}
& \textbf{17.764} & 0.475 & \textbf{0.333}
& \textbf{0.202} & \textbf{0.149} \\

\bottomrule
\end{tabular}%
}
\caption{\textbf{Ablation study.}
Component, selection-strategy, and geometry-evaluator ablations using Wan on a fixed subset of 50 held-out scenes drawn from the evaluation set used for baseline comparison with resolution $256\times256$.}
\label{tab:ablation}
\vspace{-5pt}
\end{table}
\begin{figure}[t]
    \centering
    \includegraphics[width=0.9\linewidth]{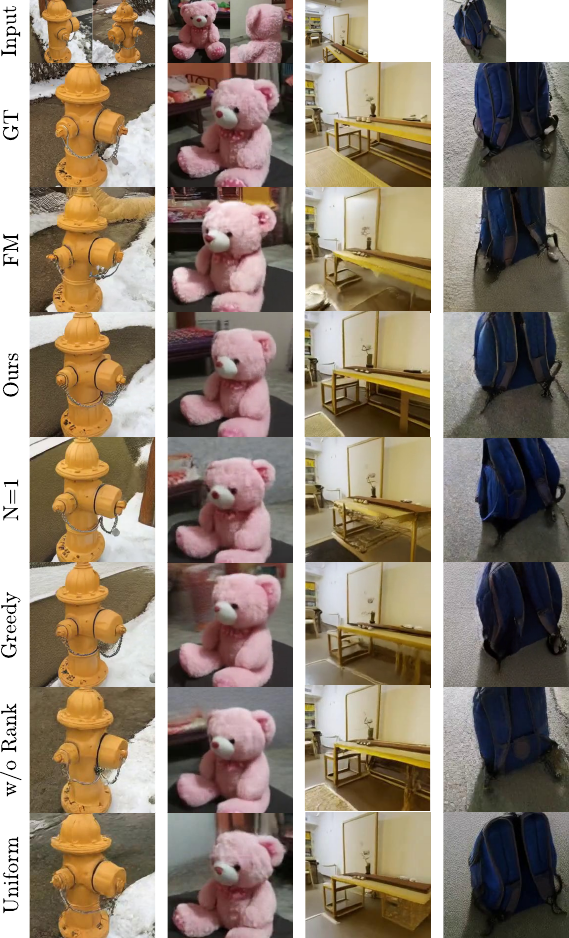}
    \caption{\textbf{Qualitative ablation study.}}
    \label{fig:ablation}
    \vspace{-3mm}
\end{figure}
\begin{figure}[t]
\centering
\includegraphics[width=0.95\linewidth]{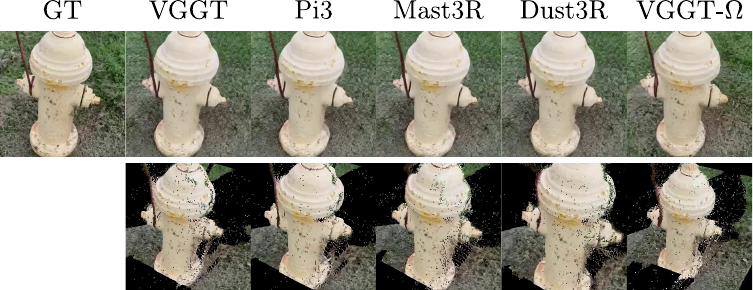}
\caption{\textbf{Effect of geometry evaluators.}
GeoART produces consistent results across different off-the-shelf geometry estimators, demonstrating robustness to the choice of geometry prior.}
\label{fig:backbone}
\vspace{-3mm}
\end{figure}

\noindent\textbf{Without candidate-space scaling.}
Setting $N=1$ retains P\&P refinement but eliminates candidate-space exploration. As shown in Table~\ref{tab:ablation} and Fig.~\ref{fig:ablation}, expanding the pool to the default $N=5$ improves all reported metrics and produces more faithful appearance and camera alignment, demonstrating the benefit of exploring multiple local refinement directions.

\noindent\textbf{Candidate selection.}
We compare our rank-calibrated probabilistic selection with uniform,
greedy, and score-only alternatives. Uniform selection ignores geometry,
whereas greedy selection always chooses the highest-scoring candidate and
is therefore sensitive to evaluation noise. Removing rank calibration
reduces the method to score-only sampling,
$p_k\propto\exp(\lambda s_k)$, whose behavior depends on the score scale.
Our full strategy achieves the strongest overall performance by combining
score magnitude with scale-invariant relative rank, favoring geometrically
consistent candidates while retaining controlled exploration(Table~\ref{tab:ablation}, Fig.~\ref{fig:ablation}).

\noindent\textbf{Different geometry scoring models.}
We evaluate GeoART using Pi3, VGGT-$\Omega$, MASt3R, DUSt3R, and VGGT.
Performance remains consistently strong across all five evaluators,
demonstrating that GeoART is compatible with different off-the-shelf
geometry priors. Although no evaluator dominates every metric, VGGT
provides the best overall balance, while the small variation across models shows that GeoART does not require perfect geometry reconstruction(Table~\ref{tab:ablation}, Fig.~\ref{fig:backbone}).

\noindent\textbf{Confidence gate.}
Removing confidence weighting causes consistent degradation across
reconstruction and pose metrics(Table~\ref{tab:ablation}). Without the gate, unreliable evidence
from occluded, textureless, or poorly reconstructed regions contributes
equally to candidate scoring and can misdirect selection. Confidence
weighting suppresses these regions, improving the robustness of
geometry-aware refinement.

\section{Conclusion}
\label{sec:conclusion}
We introduced GeoART, a training-free geometry-aware self-refinement framework that improves camera controllability and geometric consistency through test-time scaling. GeoART uses geometry only to evaluate refinement candidates, while confidence-gated scoring and rank-calibrated probabilistic selection reduce sensitivity to unreliable estimates. Experiments across two backbones demonstrate consistent gains in camera alignment, multi-view consistency, and visual quality without retraining. Future work will focus on reducing inference cost and improving geometry estimation for videos with moving objects.
\clearpage
\setcounter{page}{1}
\maketitlesupplementary

\subsection{More Implementation Details}
\label{sec:baseline_implementation}

\paragraph{Common evaluation setup.}
We evaluate all baselines using the same trained Wan2.1-Fun-V1.1-1.3B-Control-Camera~\cite{wan} and Uni3C~\cite{uni3c}
camera-conditioned video-generation checkpoints under both one-view and
two-view settings. In the one-view setting, a single input image conditions image-to-video generation. In the two-view setting, the first and last views serve as head-and-tail conditions for video interpolation. Unless otherwise specified, each generated video contains
33 frames at a spatial resolution of $256\times256$. To improve precise camera control, we fine-tune both models on combination of CO3D and DL3DV. Each ground-truth clip spans 40 source frames, from which 33 frames are randomly sampled for training. Camera poses are expressed relative to the first camera,
and per-pixel Pl\"ucker ray embeddings are computed from the corresponding
camera intrinsics and extrinsics. All methods use identical prompts, conditioning observations, camera trajectories, model checkpoints,
and random seeds. We use seed 0,10,20,30,40 and evaluate 50 videos from each
dataset. Sampling is performed in \texttt{bfloat16} using a flow-matching
scheduler with a shift parameter of 5 and tiled VAE decoding. The follwing are implementation details of other baselines. 

\paragraph{Flow-matching baselines.}
The FM baseline uses the standard sampler provided by each video
backbone. At every solver step, conventional classifier-free guidance
(CFG) combines the conditional and unconditional predictions as
\begin{equation*}
\epsilon_\theta^{\mathrm{CFG}}(z_t,t,y)
=
\epsilon_\theta(z_t,t,\varnothing)
+
w
\left[
\epsilon_\theta(z_t,t,y)
-
\epsilon_\theta(z_t,t,\varnothing)
\right]
\end{equation*}
where $w$ is the default guidance scale of the
corresponding backbone and selected as 3.0 in experiments. Standard FM uses 50 solver steps, whereas the
higher-compute variant, denoted FM ($2\times$ NFE), uses 100 steps with
all other settings unchanged. In our reporting, NFE denotes the number
of solver steps; the conditional and unconditional CFG branches are not
counted separately. Neither variant uses test-time refinement, candidate
selection, rejection sampling, or an external verifier.

\paragraph{Self-Refining Video Sampling.}
We implement the Predict-and-Perturb strategy of Self-Refining Video
Sampling follwing~\cite{selfrefine}. At selected noise levels, the
current noisy latent is first mapped to a predicted clean sample and is then
perturbed back to the same noise level using newly sampled Gaussian noise.
Repeated prediction and perturbation progressively move the latent toward
higher-density regions of the learned video manifold. For Wan and Uni3C, we perform two additional refinement iterations at sampling steps
5--15 and one additional iteration at steps 40--45. We use an $\ell_1$
prediction-difference threshold of 0.1 and terminate refinement at the current
noise level when more than $99.9\%$ of latent locations are considered
converged. This baseline does
not require an external reward model or additional training. 

\paragraph{CFG-Ctrl.}
We adapt Sliding-Mode-Control CFG (SMC-CFG) from
CFG-Ctrl~\cite{cfgctrl} to our two video-generation backbones. Let
\begin{equation}
    e_t = v_{\mathrm{cond}}(z_t,t)
          - v_{\mathrm{uncond}}(z_t,t)
\end{equation}
denote the difference between the conditional and unconditional velocity
predictions. Following a two-step conditional-only warm-up, we construct the
sliding variable
\begin{equation}
    s_t = (e_t-e_{t-1}) + \lambda e_{t-1},
\end{equation}
and modify the guidance signal using
\begin{equation}
    \widetilde{e}_t
    = e_t - K\,\operatorname{sign}(s_t).
\end{equation}
The final guided velocity is
\begin{equation}
    \widetilde{v}_t
    = v_{\mathrm{uncond}}(z_t,t)
      + w\,\widetilde{e}_t,
\end{equation}
where $w$ is the CFG scale. We use $\lambda=5.0$, $K=0.3$,
$w=5.0$, and 50 sampling steps. All generator parameters remain frozen.

\paragraph{Z-Sampling.}
We implement Zigzag Diffusion Sampling (Z-Sampling) following
\cite{zsample}. The standard denoising trajectory is
augmented with a backward--forward refinement during the first 40 of the
50 sampling steps. At each selected step, we first perform the ordinary
strongly guided denoising update. The resulting latent is then approximately
inverted toward the preceding noise level using weak guidance
$\gamma_{\mathrm{weak}}=0$, after which it is denoised again using strong
guidance $\gamma_{\mathrm{strong}}=5$. The second denoised result replaces
the original update. This gives 50 ordinary transitions and 80 additional
zigzag transitions, for a total of 130 trajectory evaluations. 

\paragraph{Video-T1.}
We implement the linear-search variant of
Video-T1~\cite{videotts}. For each input, we independently generate $N=10$
complete videos using random seeds $s$ and $s+1$, with 50 solver steps per
candidate. Each candidate is evaluated using the released
VisionReward-Video verifier. The verifier uniformly samples 24 frames from
the generated video and evaluates the released set of prompt-conditioned
binary quality questions. The binary answers are mapped to $+1$ and $-1$,
weighted using the released question weights, and averaged. We retain the
candidate with the higher prompt-conditioned weighted score. We use only
candidate-level linear search and do not employ the Tree-of-Frames search variant.

\paragraph{VIGOR.}
For VIGOR~\cite{vigor}, we use the same best-of-$N$ protocol with $N=10$,
but replace the semantic verifier with the released geometry-oriented reward.
Each candidate is evaluated using the official VGGT-based pointwise
reprojection-error evaluator, and the candidate with the lower reprojection
error is selected. We use the default released VIGOR configuration and
disable the optional sky-region filter. The generator is not fine-tuned;
VIGOR is used exclusively as an inference-time path verifier.

\paragraph{GC-Ref.}
We adapt Geometric Consistency Refinement
(GC-Ref)~\cite{bengtson2025geometric} to video by optimizing only the initial diffusion
noise while keeping the generator weights, conditioning observations, and
camera parameters frozen. We first generate an initial video and use the RoMa matcher in the paper to establish dense correspondences between
the first conditioning image and selected generated frames. The reference
pixel locations and confidence mask are fixed from this initial prediction,
while the corresponding target locations are recomputed differentiably
during optimization.

The refinement objective is
\begin{equation}
    \mathcal{L}_{\mathrm{GC\text{-}Ref}}
    =
    \mathcal{L}_{\mathrm{epi}}
    +
    \lambda_{\mathrm{RGB}}
    \mathcal{L}_{\mathrm{RGB}},
\end{equation}
where $\mathcal{L}_{\mathrm{epi}}$ is a Huber penalty on the symmetric
point-to-epipolar-line distance, and $\mathcal{L}_{\mathrm{RGB}}$ is the
$\ell_1$ color difference between matched pixels. We use five Adam
optimization iterations, a learning rate of 0.025, a RoMa confidence
threshold of 0.15, $\lambda_{\mathrm{RGB}}=2.5$, a Huber threshold of 2.0,
a temporal frame stride of 4, and at most two target frames. Each
optimization iteration performs a complete 50-step denoising trajectory with
CFG scale 3. After optimization, a final trajectory is generated from the
refined initial noise.

% Required packages: booktabs, tabularx, array
\subsection{Inference Hyperparameters}
\label{sec:infer_ablation_v16_hyperparameters}

We use the same configuration for all experiments of all methods. The default hyperparameters are provided in Table~\ref{tab:infer_hyperparameters}.

\begin{table}[t]
  \centering
  \small
  \setlength{\tabcolsep}{5pt}
  \caption{Hyperparameters of GeoART.}
  \label{tab:infer_hyperparameters}
  \begin{tabularx}{\linewidth}{@{}>{\raggedright\arraybackslash}X l@{}}
    \toprule
    Hyperparameter & Value \\
    \midrule
    Spatial resolution & $256\times256$ \\
    Number of frames & 33 \\
    Number of denoising steps & 50 \\
    Denoising strength & 1.0 \\
    Classifier-free guidance scale & 3.0 \\
    Flow-scheduler shift & 5.0 \\
    Random seed for diffusion sampling & 0,10,20,30,40 \\
    VAE tile size & $(30,52)$ \\
    VAE tile stride & $(15,26)$ \\
    Geometry estimator & VGGT \\
    Number of candidates $K$ & 5 \\
    Selection interval & 5 denoising steps \\
    Stochastic refinement schedule & $(5{:}15,2),(40{:}45,1)$ \\
    Score-softmax inverse temperature $\lambda$ & 3.0 \\
    Numerical constant $\epsilon$ & $10^{-6}$ \\
    confidence gate $\tau$ & 0.3 \\
    \bottomrule
  \end{tabularx}
\end{table}

The stochastic refinement schedule uses zero-based denoising-step indices.
The notation $(a{:}b,r)$ means that $r$ extra refinement iterations are applied
at every step in the inclusive interval $[a,b]$. Candidate search is
executed every five steps within these refinement intervals.

% Required packages: \usepackage{booktabs,graphicx}
\subsection{Hyperparameter Analysis}
\label{sec:supp_hyperparameters}

We analyze two hyperparameters of GeoART: the confidence threshold used by the
geometry score and the temporal schedule used for stochastic refinement.  All
experiments use the same seed ($0$), geometry backbone (VGGT), candidate count
($K=5$), and candidate-selection interval (every five denoising steps).  We
evaluate the same matched subset of 20 videos from CO3D and DL3DV.  Unless it is the variable being swept, every
other hyperparameter is held fixed.

\paragraph{Confidence gate.}
Let $c$ denote the confidence predicted by the geometry backbone and let
$\tau_c$ denote the confidence threshold.  In contrast to a binary mask, we
retain the original confidence magnitude above the threshold:
\begin{equation}
  w(c;\tau_c) = c\,\mathbf{1}[c\geq\tau_c].
  \label{eq:supp_soft_confidence_gate}
\end{equation}
The geometry score for candidate $k$ is therefore
\begin{equation}
  s^k = -\operatorname{mean}\!\left[
    \left(\hat z^{k}-g\right)^2
    \odot w(c;\tau_c)
  \right].
  \label{eq:supp_confidence_geometry_score}
\end{equation}
Here, $\odot$ denotes element-wise multiplication. 
We sweep
\begin{equation}
  \tau_c\in\{0.1,0.2,0.3,0.4,0.5,0.6,0.7,0.8\}
\end{equation}
while keeping the refinement schedule fixed to
$\mathcal{R}=\{(5,15,2),(40,45,1)\}$.
Table~\ref{tab:supp_confidence_threshold} shows a clear optimum at $\tau_c=0.3$, which achieves the best performance across all metrics. Lower thresholds retain unreliable geometric estimates, weakening candidate evaluation. Conversely, higher thresholds discard too many valid regions and reduce the coverage of the geometry signal, with performance degrading notably at $\tau_c\geq0.7$. These results demonstrate that the confidence gate balances geometric reliability and coverage; we therefore use $\tau_c=0.3$ by default.
\begin{table}[t]
  \centering
  \resizebox{\linewidth}{!}{%
  \begin{tabular}{crrrrrrr}
    \toprule
    $\tau_c$ & PSNR$\uparrow$ & SSIM$\uparrow$ & LPIPS$\downarrow$ &
    Trans.$\downarrow$ & Rot.$\downarrow$ & FVD$\downarrow$ &  \\
    \midrule
    0.1 & 17.17 & 0.400 & 0.316 & 0.216 & 0.187 & 103.14  \\
    0.2 & 17.64 & 0.493 & 0.296 & 0.159 & 0.118 & 76.35 \\
    0.3 &\textbf{17.76} & \textbf{0.498} & \textbf{0.275} & \textbf{0.135} & \textbf{0.096} & \textbf{75.99} \\ 
    0.4 & 17.67 & 0.490 & 0.293 & 0.149 & 0.105 & 77.79 \\
    0.5 & 17.53 & 0.475 & 0.314 & 0.172 & 0.144 & 85.22 \\
    0.6 & 17.41 & 0.430 & 0.306 & 0.206 & 0.185 & 94.36 \\
    0.7 & 17.28 & 0.423 & 0.315 & 0.236 & 0.194 & 105.46 \\
    0.8 & 17.04 & 0.393 & 0.326 & 0.249 & 0.205 & 110.14 \\
    \bottomrule
  \end{tabular}%
  }
  \caption{Confidence-threshold sensitivity.}
  \label{tab:supp_confidence_threshold}
  
\end{table}

\paragraph{Refinement schedule.}
For this analysis in Table~\ref{tab:supp_refinement_schedule}, the soft confidence threshold is fixed to $\tau_c=0.3$.
We compare early-only, late-only, and combined refinement schedules.

\begin{table}[t]
  \centering
  \resizebox{\linewidth}{!}{%
  \begin{tabular}{lrrrrrr}
    \toprule
    Refinement schedule $\mathcal{R}$
    & PSNR$\uparrow$
    & SSIM$\uparrow$
    & LPIPS$\downarrow$
    & Trans.$\downarrow$
    & Rot.$\downarrow$
    & FVD$\downarrow$ \\
    \midrule
    $\{(5,10,2)\}$
      & 17.26 & 0.481 & 0.298 & 0.201 & 0.143 & 94.73 \\
    $\{(5,15,2)\}$
      & 17.64 & 0.496 & 0.281 & 0.162 & 0.115
      & {79.74} \\

    $\{(5,20,2)\}$
      & 17.70 & 0.496 & 0.283 & 0.175 & 0.124 & 80.87 \\
    $\{(5,30,2)\}$
      & 17.64 & 0.494 & 0.286 & 0.171 & 0.122 & 81.12 \\

    $\{(40,45,1)\}$
      & 17.48 & 0.485 & 0.289 & 0.168 & 0.119 & 92.84 \\
    $\{(5,15,2),(40,45,1)\}$
      & \textbf{17.76} & \textbf{0.498} & \textbf{0.275}
      & \textbf{0.135} & \textbf{0.096} & \textbf{75.99} \\
    \bottomrule
  \end{tabular}%
  }
  \caption{Sensitivity to the stochastic-refinement schedule.}
  \label{tab:supp_refinement_schedule}
  
\end{table}

The combined schedule $\{(5,15,2),(40,45,1)\}$ gives the best reconstruction
quality and camera accuracy, suggesting that early refinement establishes
global geometric consistency while a short late-stage refinement interval
corrects residual geometric drift.  Extending early refinement to step 20 or 30 increases
computation without consistently improving quality, indicating diminishing
returns from longer refinement windows.

\subsection{Higher Resolution Result}
\begin{figure*}[t]
    \centering
    \includegraphics[width=0.6\textwidth]{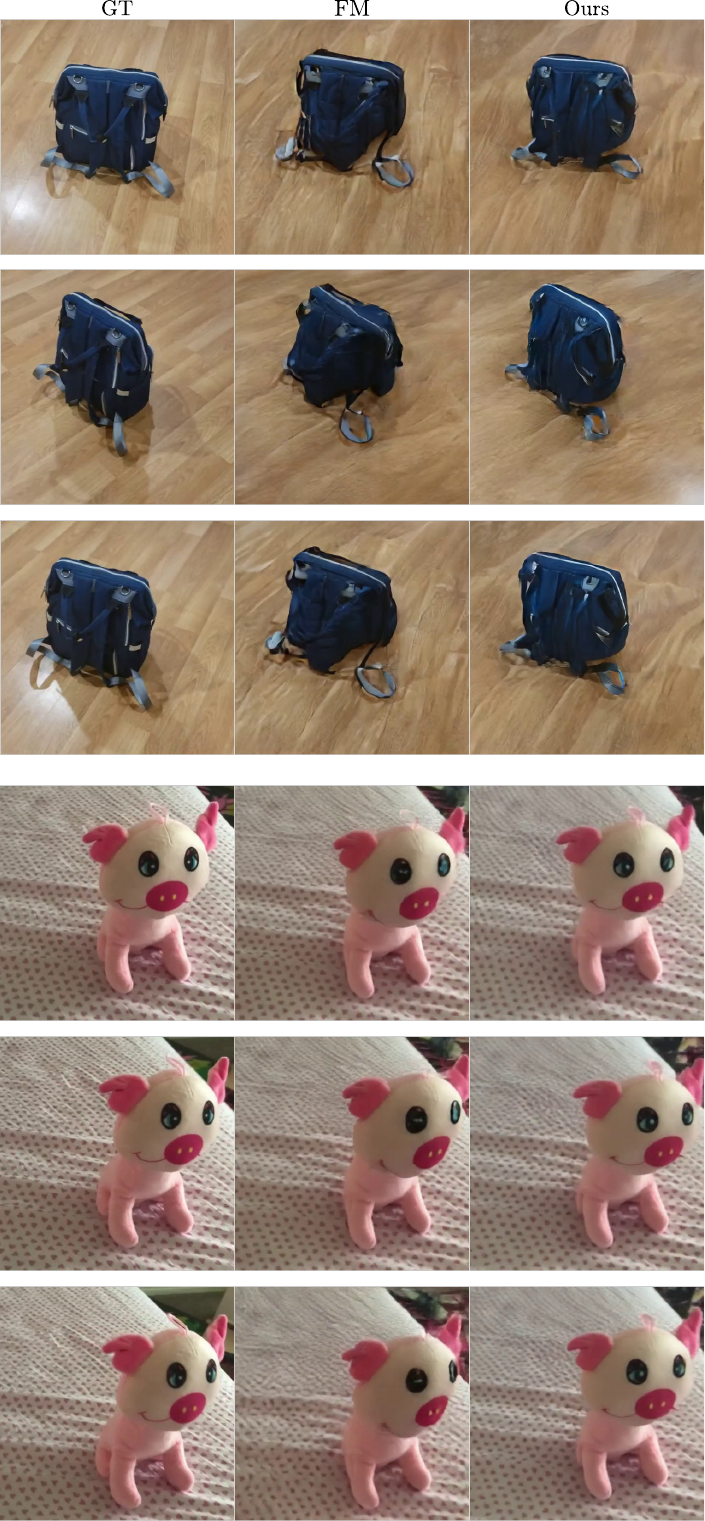}
    \caption{\textbf{Qualitative comparison at $512\times512$ resolution.}}
    \label{fig:sup-high1}
    \vspace{-3mm}
\end{figure*}

\begin{figure*}[t]
    \centering
    \includegraphics[width=0.6\textwidth]{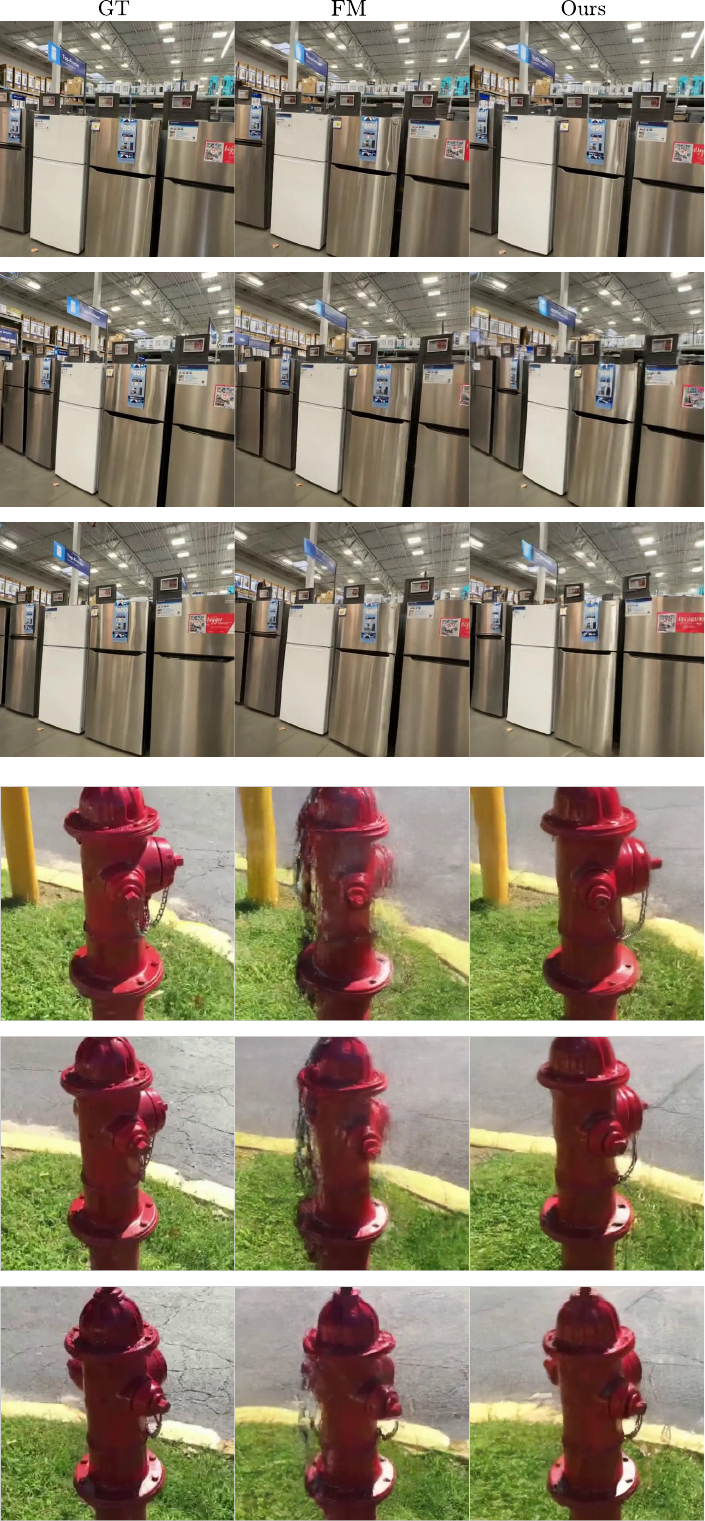}
    \caption{\textbf{Qualitative comparison at $512\times512$ resolution.}}
    \label{fig:sup-high2}
    \vspace{-3mm}
\end{figure*}

We additionally evaluate Wan two-view generation at an inference resolution of
$512\times512$, four times the number of pixels used in our $256\times256$
evaluation.  We use the same model checkpoints and compare the baseline methods with GeoART (Ours, $K=5$).  Both methods use seed 40 and the
same input views and camera trajectories. The quantatitive results are provided in Table~\ref{tab:supp_512_quantitative}. The visual results are provided in Figure~\ref{fig:sup-high1},~\ref{fig:sup-high2}.

\begin{table}[t]
\centering
\setlength{\tabcolsep}{3pt}
\renewcommand{\arraystretch}{1.06}
\scriptsize
\resizebox{\columnwidth}{!}{%
\begin{tabular}{@{}lccccc@{}}
\toprule
\textbf{Method}
& \textbf{PSNR} $\uparrow$
& \textbf{SSIM} $\uparrow$
& \textbf{LPIPS} $\downarrow$
& \textbf{Trans.} $\downarrow$
& \textbf{Rot.} $\downarrow$ \\
\midrule
FM
& 16.69 & 0.386 & 0.434 & 0.289 & 0.243 \\
Self-Refine~\cite{selfrefine}
& 17.24 & 0.394 & 0.426 & 0.256 & 0.204 \\
CFG-Ctrl~\cite{cfgctrl}
& 16.88 & 0.388 & 0.443 & 0.284 & 0.253 \\
Video-T1~\cite{videotts}
& 17.13 & 0.392 & 0.412  &0.221 & 0.183 \\
\textbf{Ours}
& \textbf{17.83} & \textbf{0.439} &\textbf{ 0.387} & \textbf{0.185} & \textbf{0.144} \\
\bottomrule
\end{tabular}%
}
\caption{\textbf{Higher-resolution evaluation.}
Comparison at $512\times512$ resolution on held-out sequences.}
\label{tab:supp_512_quantitative}
\vspace{-2mm}
\end{table}

\subsection{Additional Qualitative Results}
\label{sec}

\begin{figure*}[t]
    \centering
    \includegraphics[width=0.9\textwidth]{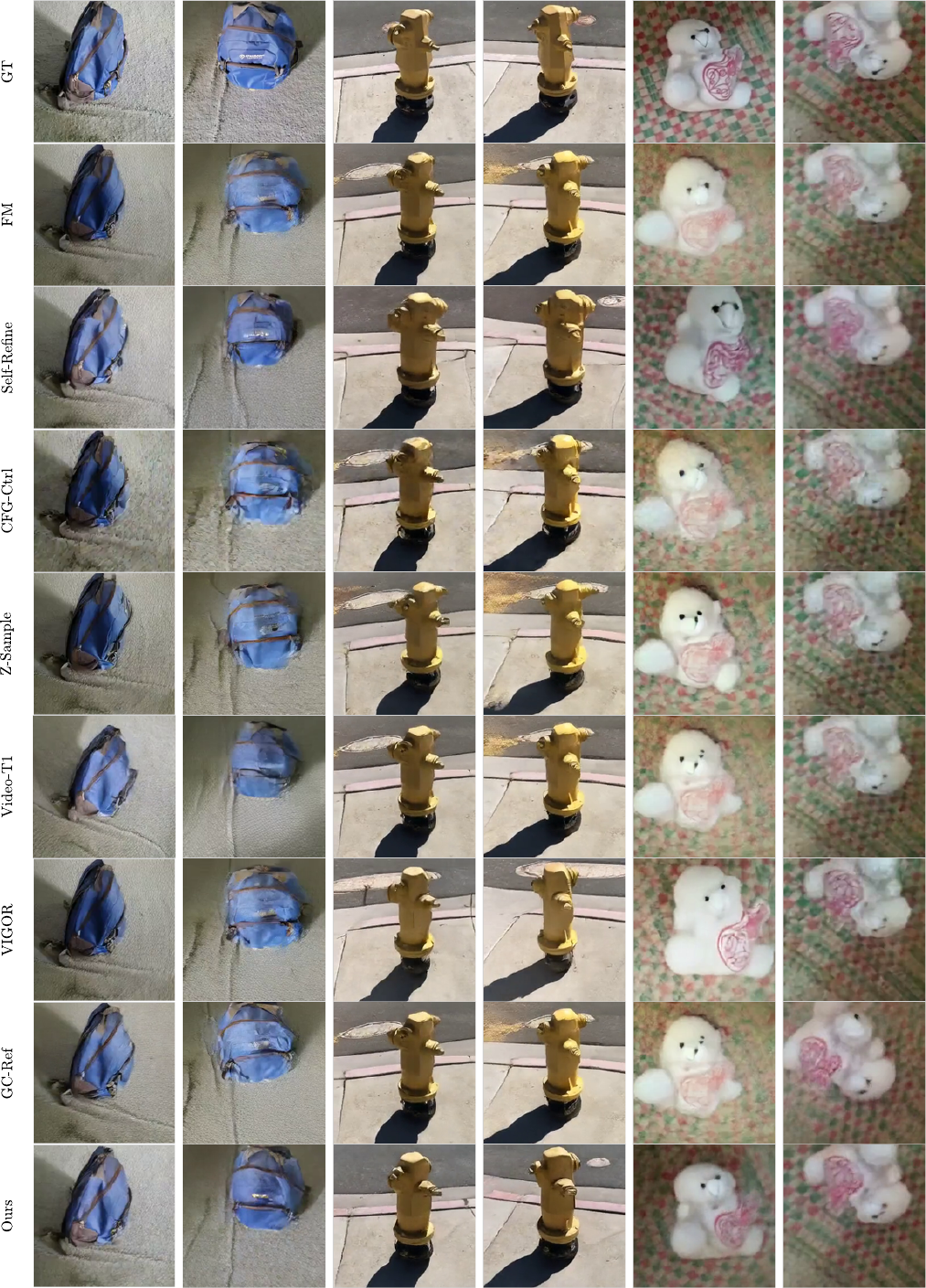}
    \caption{\textbf{Comparison with baseline methods.}}
    \label{fig:sup-baseline-comparison1}
    \vspace{-3mm}
\end{figure*}

\begin{figure*}[t]
    \centering
    \includegraphics[width=0.9\textwidth]{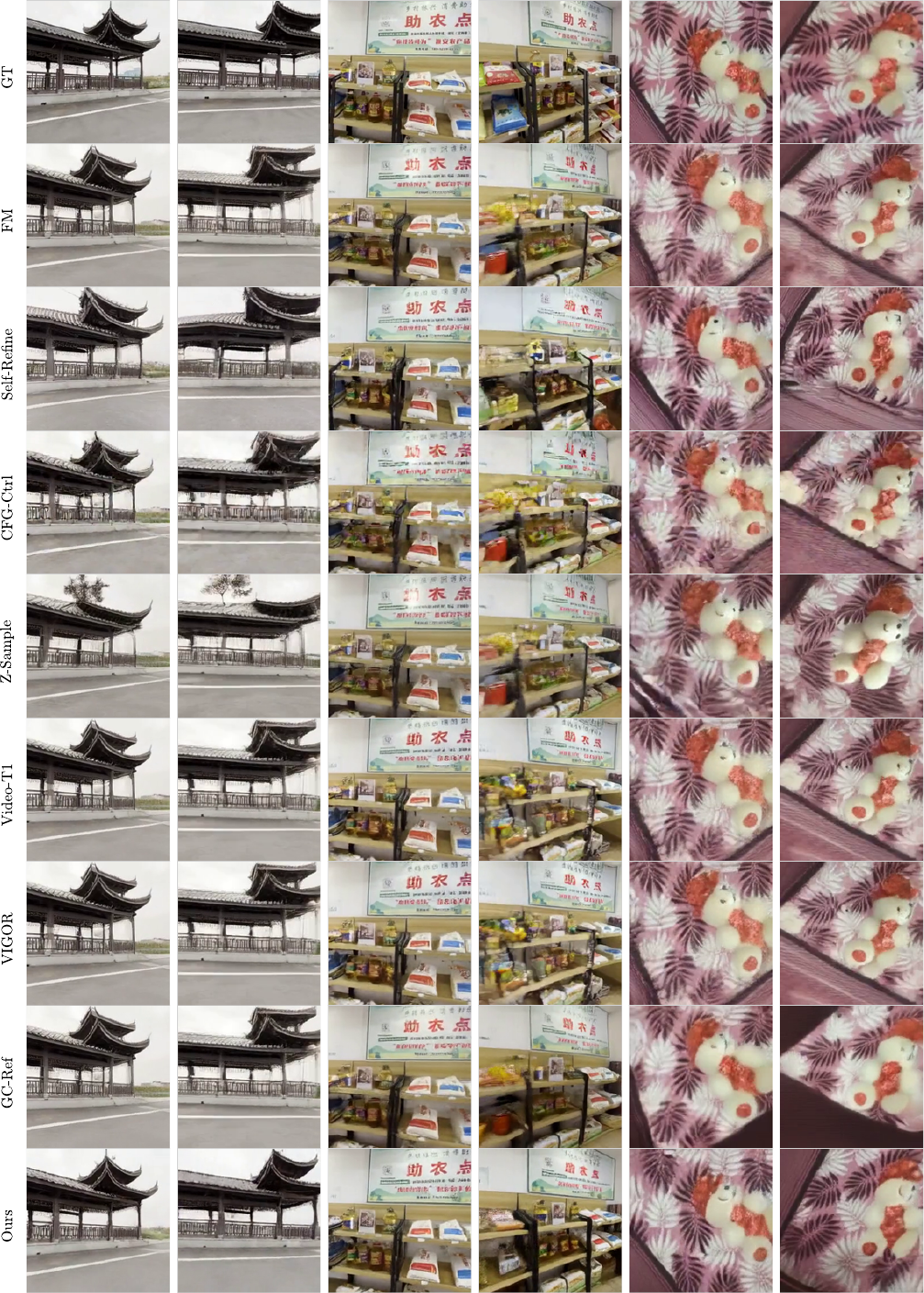}
    \caption{\textbf{Comparison with baseline methods.}}
    \label{fig:sup-baseline-comparison2}
    \vspace{-3mm}
\end{figure*}

Figure~\ref{fig:sup-baseline-comparison1} and \ref{fig:sup-baseline-comparison2} provide additional qualitative comparisons with baseline methods. Across different scenes and camera trajectories, GeoART produces more coherent novel views with better camera alignment and fewer structural artifacts. In contrast, the baseline methods more frequently exhibit positional drift, geometric deformation, and cross-frame inconsistencies. These results further demonstrate the effectiveness of geometry-aware self-refinement across diverse inputs.

\subsection{Failure Cases}
\label{sec:failure_cases}

GeoART has two main failure modes. First, it inherits errors from the geometry estimator. In textureless, reflective, transparent, or heavily occluded regions, VGGT may produce inaccurate geometry or camera correspondences. Although confidence gating suppresses uncertain predictions, it cannot remove errors assigned high confidence. Such errors can bias candidate selection and cause warped structures, duplicated details, or temporal flickering.

\begin{figure}[t]
    \centering
    \includegraphics[width=0.9\linewidth]{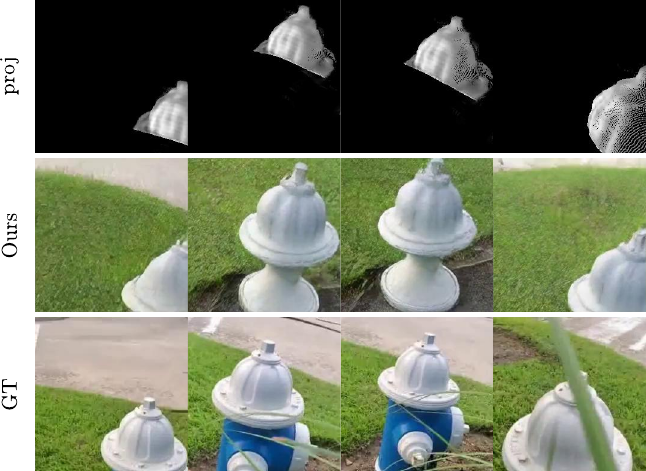}
    \caption{\textbf{Failure Case.}}
    \label{fig:fail}
    \vspace{-3mm}
\end{figure}

Second, GeoART searches over a finite, randomly generated candidate pool. With $N=5$, the pool may not contain a suitable update at a difficult denoising step. Consequently, GeoART may preserve the overall geometry and camera alignment while sacrificing fine texture quality. Together, inaccurate geometry estimates and limited candidate coverage account for the primary failure cases of our method(Figure~\ref{fig:fail}).
\subsection{Limitations}
\label{sec:limitations}

The proposed method depends on a pretrained geometry estimator and assumes
that camera intrinsics and extrinsics are available or can be estimated
reliably. Its performance is consequently bounded by the accuracy and domain generalization of VGGT. The confidence threshold is also selected heuristically,
and a fixed threshold may not transfer optimally to scenes with different
image statistics, motion patterns, or camera baselines.

Geometry-guided test-time scaling increases inference cost relative to standard diffusion
sampling. These operations increase runtime and memory consumption, limiting
the practicality of the current implementation for extremely long or high-resolution
videos.

Future work could replace dataset-specific thresholds with learnable uncertainty
calibration, incorporate temporal memory and dynamic-scene geometry, and use
adaptive candidate allocation so that additional computation is concentrated
only at uncertain denoising steps.

\clearpage

{
    \small
    \bibliographystyle{ieeenat_fullname}
    \bibliography{main}
}

\end{document}